\documentclass{article} 
\usepackage{iclr2026_conference,times}

\usepackage{amsmath,amsfonts,bm}

\def\eqref#1{equation~\ref{#1}}

\def\1{\bm{1}}

\DeclareMathAlphabet{\mathsfit}{\encodingdefault}{\sfdefault}{m}{sl}
\SetMathAlphabet{\mathsfit}{bold}{\encodingdefault}{\sfdefault}{bx}{n}

\usepackage{amsmath}
\usepackage{amssymb}
\usepackage{amsthm}
\usepackage{booktabs}
\usepackage{enumitem}
\usepackage{graphicx}
\usepackage{makecell}
\usepackage{mathtools}
\usepackage{marvosym}
\usepackage{microtype}
\usepackage{multicol}
\usepackage{multirow}
\usepackage{ragged2e}
\usepackage{subcaption}
\usepackage{tabularx}
\usepackage[table,dvipsnames]{xcolor}
\usepackage{tikz}
\usepackage{url}
\usepackage{wrapfig}
\usepackage{xspace}

\usepackage[pagebackref=false,colorlinks,citecolor=magenta,urlcolor=magenta]{hyperref}

\usepackage[capitalize]{cleveref}
\crefname{section}{Sec.}{Secs.}
\Crefname{section}{Section}{Sections}
\Crefname{table}{Table}{Tables}
\crefname{table}{Tab.}{Tabs.}

\definecolor{forestgreen}{rgb}{0.133, 0.545, 0.133}
\definecolor{yellowyellow}{rgb}{0.133, 0.545, 0.133}

\definecolor{correct}{RGB}{173, 173, 173}
\definecolor{incorrect}{RGB}{192, 0, 0}

\newcommand{\method}{\textsc{TIGaussian}\xspace}\usepackage{xspace}

\title{\method: Disentangle Gaussians for Spatial-Awared Text-Image-3D Alignment}

\author{
Jiarun Liu\textsuperscript{1}\thanks{Equal contribution.}\quad
Qifeng Chen\textsuperscript{1}\footnotemark[1]\quad
Yiru Zhao\textsuperscript{1}\quad
Minghua Liu\textsuperscript{2}\quad
Baorui Ma\textsuperscript{3}\quad
Sheng Yang\textsuperscript{1}\thanks{Corresponding author.}
\\
\textsuperscript{1}Unmanned Vehicle Dept., Cainiao Inc., Alibaba Group, Hangzhou, China \\
\textsuperscript{2}Hillbot, Sunnyvale, USA \\
\textsuperscript{3}Beijing Academy of Artificial Intelligence, Beijing, China \\
\texttt{jiarunliu@zju.edu.cn, \{cqf7419, shengyang93fs\}@gmail.com}
}

\iclrfinalcopy 
\begin{document}

\maketitle

\begin{figure*}[ht]
    \centering
    \includegraphics[width=\linewidth]{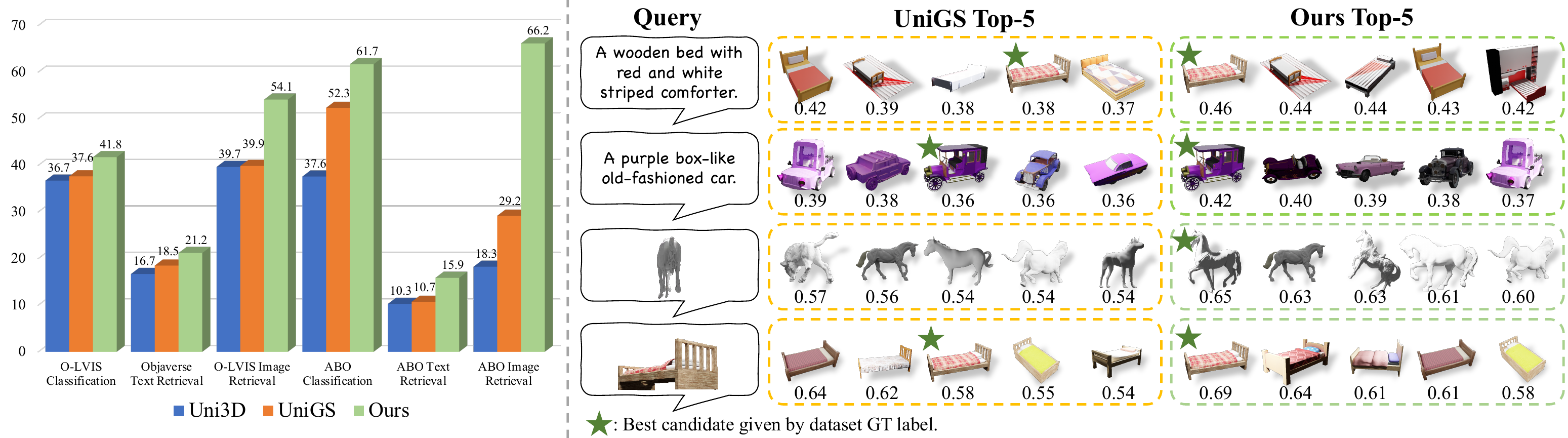}
    \caption{\method enables 3D modal pretraining on several tasks, e.g., zero-shot classification, text-3D retrieval and image-3D retrieval. \textbf{Left:} Compared to prior 3D multi-modal alignment methods -- Uni3D and UniGS, \method presents superior performance on multiple datasets -- Objaverse(-LVIS) and ABO. \textbf{Right:} In challenging scenarios involving ambiguous or complex queries, \method demonstrates superior performance owing to its disentangled encoder architecture and specialized cross-modal alignment mechanism. We report the similarity score of each item in the figure.} 
    \label{fig:teaser}
\end{figure*}

\begin{abstract}
While visual-language models have profoundly linked features between texts and images, the incorporation of 3D modality data, such as point clouds and 3D Gaussians, further enables pretraining for 3D-related tasks, e.g., cross-modal retrieval, zero-shot classification, and scene recognition.
As challenges remain in extracting 3D modal features and bridging the gap between different modalities,
we propose \method, a framework that harnesses 3D Gaussian Splatting (3DGS) characteristics to strengthen cross-modality alignment through multi-branch 3DGS tokenizer and modality-specific 3D feature alignment strategies. Specifically, our multi-branch 3DGS tokenizer decouples the intrinsic properties of 3DGS structures into compact latent representations, enabling more generalizable feature extraction. To further bridge the modality gap, we develop a bidirectional cross-modal alignment strategies: a multi-view feature fusion mechanism that leverages diffusion priors to resolve perspective ambiguity in image-3D alignment, while a text-3D projection module adaptively maps 3D features to text embedding space for better text-3D alignment.
Extensive experiments on various datasets demonstrate the state-of-the-art performance of \method in multiple tasks.
Code repository: \href{https://github.com/RUiN-jiarun/TIGaussian}{https://github.com/RUiN-jiarun/TIGaussian}.
\end{abstract}


\section{Introduction}
\label{sec:intro}


Recent breakthroughs in vision-language pretraining ~\citep{radford2021learningclip,sun2023evaclip} have established text-image alignment as the foundation for multi-modal understanding. In recent years, significant advances have been made in expanding the field of 3D visual understanding through various 3D representations. Early works in 3D multi-modal alignment primarily relied on \textit{point clouds} processed through 3D convolutions~\citep{zhang2022pointclip, huang2023clip2point, xue2023ulip, xue2024ulip2} or 3D Vision Transformers ~\citep{zeng2023clip2}. Subsequently, works on \textit{meshes}~\citep{song2023meshclip} and \textit{voxels}~\citep{ruan2024tricolo} further explores the operations on organized 3D representations. More recently, UniGS~\citep{li2025unigs} first leverages \textit{3D Gaussian Splatting} (3DGS) as an advanced 3D representation, demonstrating state-of-the-art text-image-3D alignment across multiple benchmarks.

However, existing methods still facing challenges regarding the abstraction ability of 3D context, as well as the gaps between \textit{text-3D} and \textit{image-3D} modalities.
Specifically, our analysis reveals that the current 3DGS-based multi-modal work UniGS, while groundbreaking, suffers from following limitations: 
(1) \textit{Entangled 3D encoding}: All attributes of each Gaussian primitive are concatenated and encoded as a whole, insufficiently leveraged the distribution pattern and geometric significance of each attribute intrinsic relations between attributes;
(2) \textit{Degraded 3D perception}: Forced alignment from a single-view image fails to capture global context, causing a decrease in 3D perception ability.

To overcome these limitations, we propose \method, a tri-modal alignment framework that bridges text, image, and 3DGS representations which outperforms existing feature alignment approaches. 
At its core, our method utilizes multi-branch 3DGS tokenizer to extract 3D contextual information, which reducing inter-attribute coupling, ultimately establishing a compact and effective latent representation for 3DGS. 
For image-3D alignment, we incorporate a diffusion-enhanced multi-view fusion strategy, which compensates for single-view limitations by leveraging implicit 3D awareness from pretrained diffusion priors. Meanwhile, for text-3D alignment, the encoded feature adapts the 3D latent space through a 3D-text projection module, thereby better matching text embedding structures.

We conduct extensive experiments to validate the performance of \method on multiple tasks across the Objaverse~\citep{deitke2023objaverse}, ABO~\citep{collins2022abo} and SUN RGBD~\citep{song2015sun} datasets. As presents in Fig.~\ref{fig:teaser}, the results demonstrate the effectiveness of the proposed method, achieving superior performances on zero-shot classification, cross-modal retrieval, few-shot linear probing and open-world scene recognition.

In summary, our contributions can be summarized as follows:
\begin{itemize}
    \item We propose \method, a framework for text-image-3DGS alignment, achieving state-of-the-art performance on various downstream tasks.
    \item We design a multi-branch 3DGS tokenizer, with specialized Gaussian attribute modeling to enhance 3D feature abstraction and compression.
    \item We propose a tri-modal alignment strategy through 3D-aware image feature fusion and 3D-text projection module, enabling robust alignment for both text and image modalities.

\end{itemize}


\section{Related Work}
\label{sec:rel}

\subsection{3D Representation Learning. }
The development of 3D representations is an important factor in promoting research on 3D shape analysis and multi-modality tasks. Early works focused on volumetric grids~\citep{maturana2015voxnet}, which enabled structured 3D convolutions but incurred prohibitive computational costs for high-resolution scenes. With the development of neural networks, more works use point clouds as sparse geometric proxies, leveraging permutation-invariant operators like PointNet~\citep{qi2017pointnet, qi2017pointnet++} and other classical network architectures~\citep{choy20194d, ma2022pointmlp} to extract localized features. In recent years, there have been many mesh-based feature extraction works in dense 3D representation~\citep{feng2019meshnet, hu2022subdivision}, but the inherent complexity of mesh structures poses significant challenges for effective feature representation.
A transformative leap arrived with implicit neural representations like DeepSDF ~\citep{park2019deepsdf} and NeRF ~\citep{mildenhall2021nerf}, which encode surfaces or radiance fields via coordinate-based MLPs. These methods achieved unprecedented reconstruction quality but suffered from slow optimization and rendering. The recent Gaussian Splatting~\citep{kerbl3Dgaussians, huang20242dgs} further addresses these limitations by introducing explicit, differentiable primitives optimized for real-time photorealistic rendering. Some recent works exploit the hierarchical structures for scene organization and neural anchors representations~\citep{lu2024scaffold, ren2024octree} for better feature extraction, or embedding the semantic or visual features into 3D Gaussian primitives~\citep{zhou2024feature, shi2024language, qin2024langsplat}. These works in 3D representation learning in recent years have demonstrated the superiority of 3DGS in scene representation, understanding, and other aspects.

\subsection{Multi-modal Pretraining in 3D Tasks. }
\paragraph{Text-2D Multi-modal Learning.}
Multi-modal pretraining via contrastive learning has revolutionized cross-modal understanding, with foundational frameworks like CLIP ~\citep{radford2021learningclip} and EVA-CLIP~\citep{sun2023evaclip} demonstrating the power of aligning image-text pairs. BLIP series~\citep{li2022blip, li2023blip2} further enhance the text-image alignment capability through richer data and hybrid encoder-decoder structures. More recent works involve Large Language Models (LLMs) for visual-text understanding~\citep{team2023internlm, zhang2023internlm, dong2024internlm, chen2024internvl}.

\paragraph{Text-image-3D Tri-modal Learning.}
Recently, more works have extended to 3D fields. Early 3D adaptation strategies such as PointCLIP~\citep{zhang2022pointclip} and CLIP2Point~\citep{huang2023clip2point} circumvented geometric complexity by rendering depth maps from point clouds, effectively projecting 3D data into 2D subspaces for compatibility with pretrained image encoders. Recent efforts like ULIP series~\citep{xue2023ulip, xue2024ulip2} and CLIP$^2$~\citep{zeng2023clip2} directly encode raw point clouds through 3D transformers, achieving improved robustness in real-world indoor/outdoor benchmarks. Other works such as TriCoLo~\citep{ruan2024tricolo} explore the performance of voxel representations in multi-modal alignment, but are limited to the scale of data and models. OpenShape~\citep{liu2023openshape} leverages Point-BERT~\citep{yu2022pointbert} and a 3D convolution architecture to train a point cloud encoder for cross-modal understanding. The concurrent work Uni3D~\citep{zhou2023uni3d} utilizes the Vision Transformer architecture~\citep{dosovitskiy2020vit} and large amounts of point cloud data to achieve a model with 1B parameters large model, and realize a state-of-the-art solution for text-image-3D alignment. More recent works~\citep{wang2024omnibind} have adapted Uni3D as the foundation model for 3D learning. To expand into more diverse 3D representations, UniGS~\citep{li2025unigs} pioneers 3DGS for multi-modal tasks by distilling the Uni3D pretrained model, which makes significant progress. However, these methods suffer from entangled attribute encoding and single-view bias during alignment. Although recent works attempt to encode 3D objects using multi-view images~\citep{lee2024duoduo, zhou2025cross}, this inevitably increases the amount of content required for encoding and loses the simplicity of encoding directly for 3D models. Our framework addresses these limitations through an attribute-disentangled 3DGS encoding scheme coupled with cross-view attention mechanisms, dynamically aligned via a pre-projection module that warps the latent 3D manifold to match text-image contrastive embedding spaces.

\section{Methodology}

\subsection{Overview}
\begin{figure}[t]
    \centering
    \includegraphics[width=0.85\linewidth]{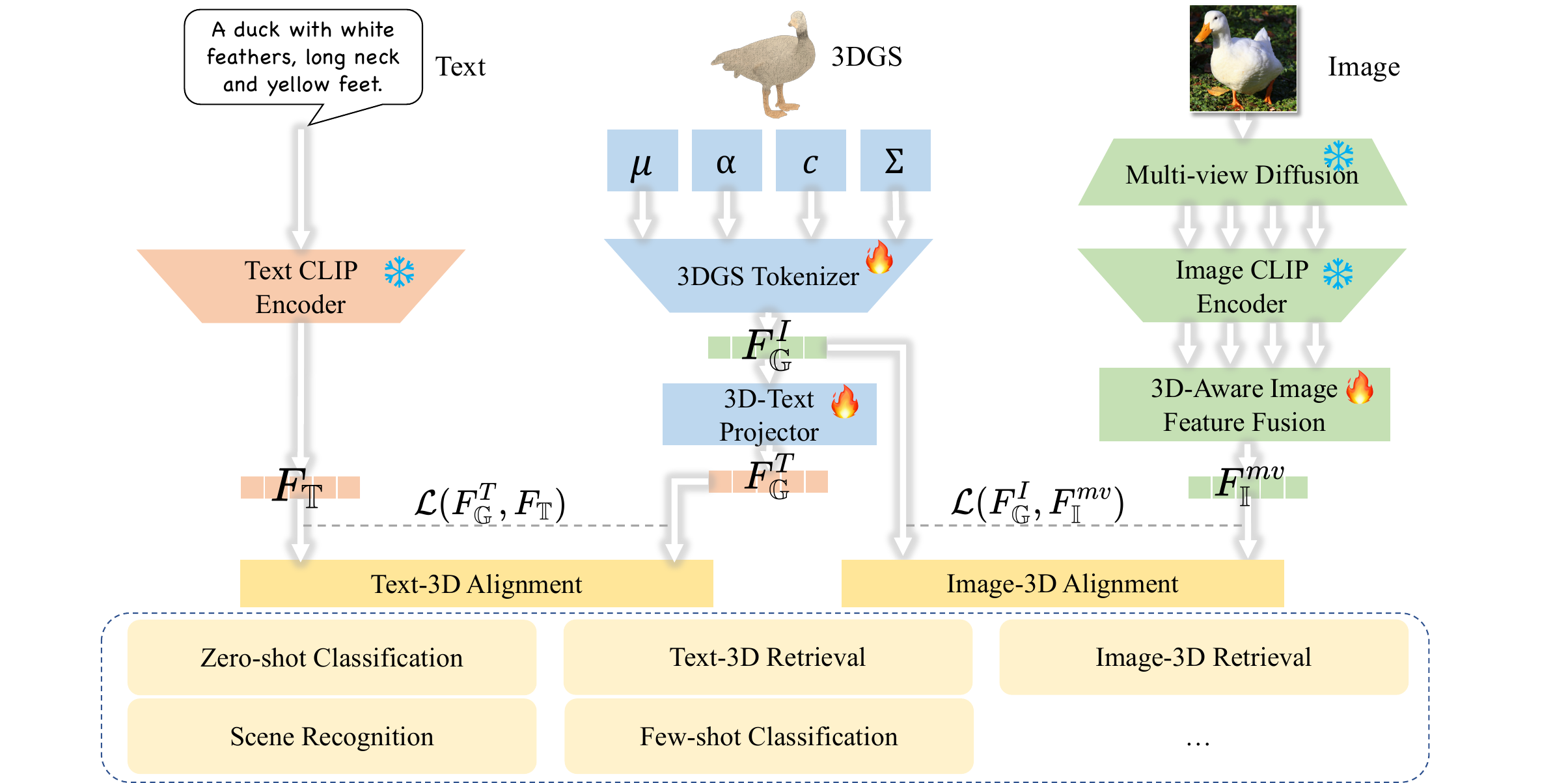}
    \caption{The \method framework for text-image-3D tri-modal representation learning. (1) \textbf{3DGS Tokenizer}: A multi-branch lightweight network decomposes input Gaussians into separate attributes, generating structured latent embedding $F_\mathbb{G}^I$. (2) \textbf{Image Modality}: Multi-view diffusion generates consistent views processed through CLIP to produce 3D-aware visual features $F_\mathbb{I}^{mv}$. (3) \textbf{Text Modality}: The 3D features are projected to text space $F_\mathbb{G}^T$ via a learnable projector. Dual contrastive losses $\mathcal{L}(F_\mathbb{G}^I, F_\mathbb{I}^{mv})$ and $\mathcal{L}(F_\mathbb{G}^T, F_\mathbb{T})$ align all modalities in a shared embedding space, enabling diverse cross-modal applications.}
    \label{fig:overview}
\end{figure}

As illustrated in Fig.~\ref{fig:overview}, with the vanilla 3DGS~\citep{kerbl3Dgaussians} as the most prevalent input Gaussian representation, \method processes \textbf{tri-modal} inputs through three coordinated components: (1) For the \textbf{3D} modal, we use a multi-branch 3DGS tokenizer which decomposes geometric and appearance attributes to construct a structured latent embedding $F_\mathbb{G}^I$ for spatial context; (2) For the \textbf{image} modal, we use a diffusion-enhanced multi-view fusion module for generating 3D-aware visual features $F_\mathbb{I}^{mv}$ which aggregates implicit 3D priors from pretrained diffusion models; (3) For aligning the \textbf{text} modal, we further project the spatial features $F_\mathbb{G}^I$ into $F_\mathbb{G}^T$ through a 3D-text projector to ensure modality-invariant feature consistency with text embeddings.

\subsection{Preliminaries}
\noindent\textbf{3D Gaussian Splatting.} 3DGS characterizes the object through a collection of anisotropic Gaussian primitives defined within the 3D space~\citep{kerbl3Dgaussians}. For each 3D Gaussian $\mathcal{G}$, it is described by the following attributes: a center position vector $\boldsymbol \mu\in \mathbb{R}^3$, an opacity parameter $\alpha \in [0,1]$, a set of spherical harmonics (SH) coefficients $\textrm{SH}\in\mathbb{R}^k$ and a covariance matrix $\boldsymbol\Sigma\in\mathbb{R}^{3\times 3}$:
\begin{equation}
    \mathcal{G}(\mathbf{x}) = \exp(-\frac{1}{2}(\mathbf{x} - \boldsymbol{\mu})^{\top} \mathbf{\Sigma}^{-1}(\mathbf{x} - \boldsymbol{\mu})),
\end{equation}
among which, the SH coefficients can be derived to RGB color $\boldsymbol c\in\mathbb{R}^3$, while the covariance $\boldsymbol\Sigma$ can be decomposed into scaling factor $\boldsymbol s\in\mathbb{R}^{3}$ and rotation quaternion parameter $\boldsymbol q\in\mathbb{R}^{4}$ as:
\begin{equation}
     \boldsymbol \Sigma=\mathbf{R}\mathbf{S}\mathbf S^\top\mathbf R^\top,
\end{equation}
where $\mathbf S\in\mathbb{R}^{3\times 3}$ is the scaling matrix derived from $\boldsymbol s$ and $\mathbf R\in\mathbb{R}^{3\times 3}$ is the rotation matrix derived from $\boldsymbol q$. 

\noindent\textbf{Multi-modal Alignment via Contrastive Learning.} After widespread validation in multi-modal works in recent years~\citep{radford2021learningclip, li2022blip, xue2023ulip}, cross-modal feature alignment through contrastive learning is a highly effective training method.  The contrastive loss of two features $\mathcal{L}(F_1, F_2)$ is calculated by the InfoNCE loss~\citep{oord2018representation}, as:
\begin{equation}
\begin{aligned}
    \mathcal{L}(F_1,F_2)&=-\mathbb{E}_{{x}}\bigg(\log\frac{\mathcal{L}^{+}(x,\tau)}{\mathcal{L}^{+}(x,\tau) + \sum_{j=1}^M \mathcal{L}^{-}(x,\tau)}\bigg), \\
    \mathcal{L}^{\otimes}(x,\tau) &\triangleq \exp(F_1(x)\cdot F_2(x^\otimes) / \tau), \quad  \otimes \in \{{+}, {-}\},
\end{aligned}
\end{equation}
where $(x,x^{+})$ denotes the positive pair of a sample $x$, and $M$ is the number of the negative pairs $\{(x,x^{-})\}_{j=1}^M$. $\tau$ is the learnable temperature parameter scaling the similarity scores in contrastive learning.
As proved in most 3D multi-modal tasks~\citep{zhou2023uni3d, li2025unigs}, the text-image model is pre-aligned~\citep{dosovitskiy2020vit, sun2023evaclip}, thus we only need to separately calculate the contrastive loss between 3D features $F_{3\mathbb{D}}$ and pre-aligned image features $F_\mathbb{I}$ and text features $F_\mathbb{T}$. Therefore, the optimization object can be described as:
\begin{equation}
    \mathcal{L}=\lambda_\mathbb{T}\mathcal{L}(F_{3\mathbb{D}},F_\mathbb{T})+\lambda_\mathbb{I}\mathcal{L}(F_{3\mathbb{D}},F_\mathbb{I}),
\label{eq:loss}
\end{equation}
where $\lambda_\mathbb{T}, \lambda_\mathbb{I}$ are balance factors.

\subsection{Multi-branch 3DGS Tokenizer}

\begin{figure}[t]
    \centering
    \begin{minipage}[t]{0.49\textwidth}
        \centering
        \includegraphics[width=\linewidth]{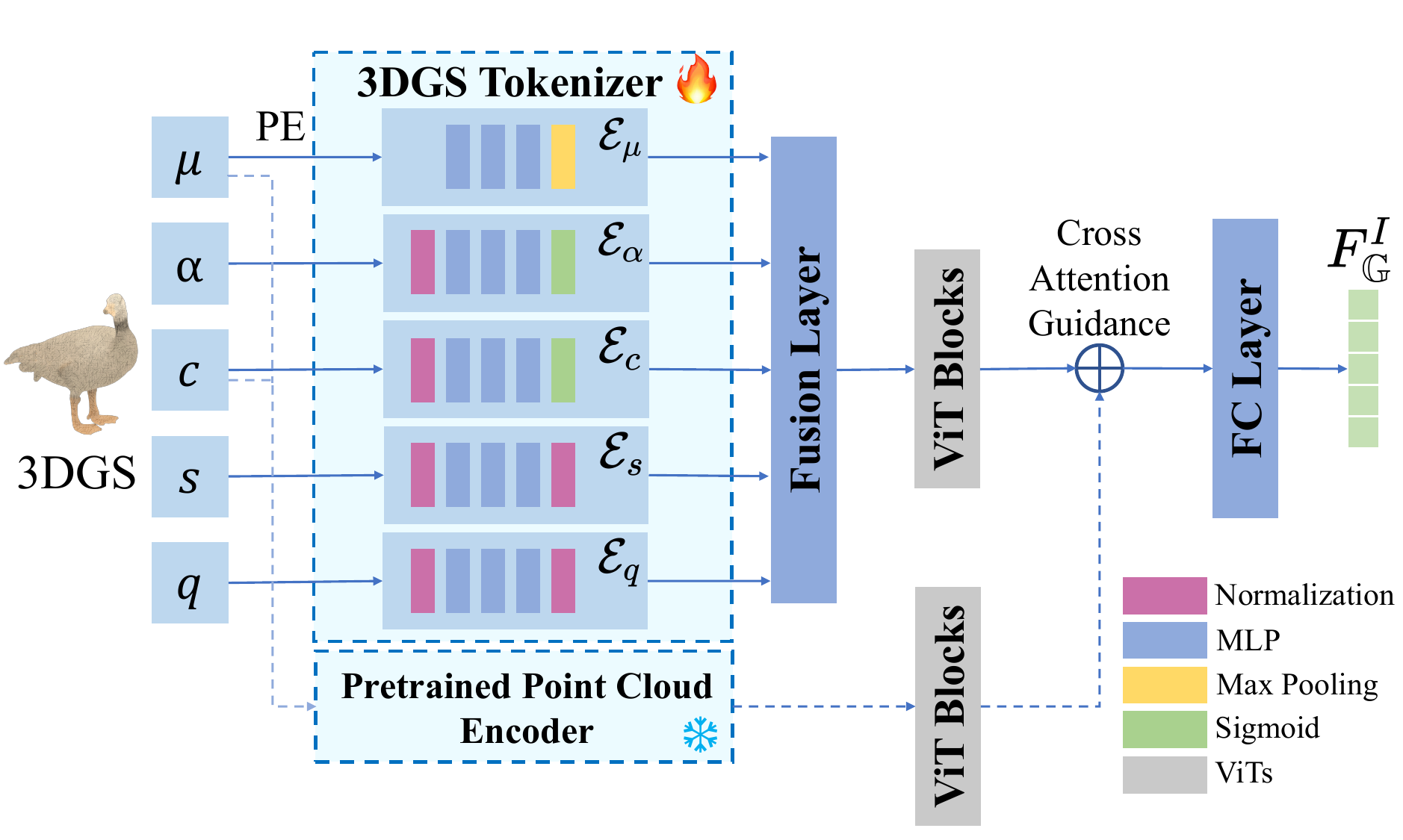}
        \caption{Illustration of 3DGS tokenizer, where multiple branches are tailored for different attributes for a more compact and effective extraction.}
        \label{fig:encoder}
    \end{minipage}
    \hfill
    \begin{minipage}[t]{0.49\textwidth}
        \centering
        \includegraphics[width=\linewidth]{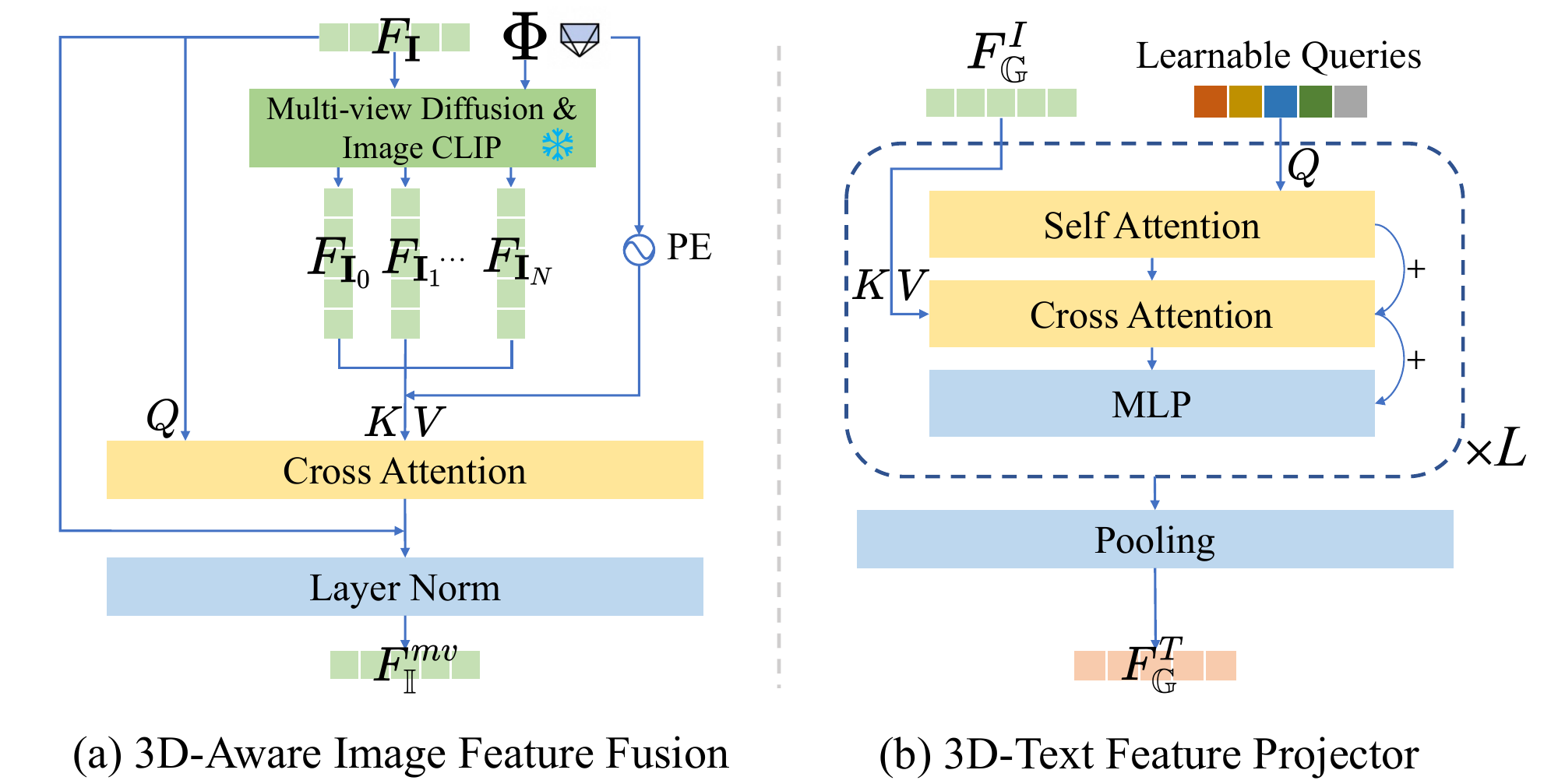}
        \caption{Illustration of 3D-aware image feature fusion module and 3D-text projector module.}
        \label{fig:align}
    \end{minipage}
\end{figure}

Given a 3D object represented by Gaussians, we first follow existing approaches~\citep{zhou2023uni3d,li2025unigs} to downsample it to 1024 Gaussians by Farthest Point Sampling (FPS) and group them by k-Nearest Neighbor (kNN) algorithm to form several local Gaussian patches of fixed numbers of Gaussians. These local groups serve as processing units for the final ViT-based encoding stage, allowing the model to capture both fine-grained local patterns and global structural relationships.
Different from UniGS~\citep{li2025unigs} which concatenates all Gaussian attributes as homogeneous features, we utilize a well-designed tokenzier to explicitly model the distinct characterization and distributional properties of Gaussian. Specifically, as shown in Fig.~\ref{fig:encoder}, each attribute of a Gaussian patch $\boldsymbol{{G}}=\{\boldsymbol \mu,\boldsymbol\alpha,\boldsymbol c,\boldsymbol s,\boldsymbol q\}$ -- including spatial ($\boldsymbol \mu$), appearance ($\boldsymbol\alpha,\boldsymbol c$) and morphological ($\boldsymbol s,\boldsymbol q$) components with inherent inter-attribute relationships -- is separately fed into its corresponding encoding branch $\{\mathcal{E}_\mu,\mathcal{E}_\alpha,\mathcal{E}_c,\mathcal{E}_s,\mathcal{E}_q\}$.
From an information representation perspective~\citep{tishby2000information}, such a disentangled token scheme reduces mutual interference among heterogeneous attributes during abstraction and compression. Each branch can adaptively compress its input under an attribute-specific information bottleneck, preserving only the most task-relevant signals.
This avoids entangled representations that arise when disparate modalities (e.g., position and color, which possess distinct numerical ranges and representational purposes) are forced to be transformed within a shared domain, thereby resulting in information loss and suboptimal feature alignment. As shown in the example of the bed in \cref{fig:teaser}, previous methods may ignore the detailed information of ``white striped comforter", while our method can highlight this feature.

Each encode branch employs a three-layer MLP for token extraction. For \textbf{spatial tokenizer} $\mathcal{E}_\mu$,
we adopt an encoder architecture inspired by those used for point clouds~\citep{qi2017pointnet}, incorporating an additional learnable positional embedding module to enhance the extraction of high-dimensional spatial context. Furthermore, a max pooling layer aggregates all point-wise features into a global descriptor by taking the channel-wise maximum values, ensuring permutation invariance and capturing the most salient geometric patterns.
For \textbf{appearance tokenizer} $\mathcal{E}_\alpha,\mathcal{E}_c$, we employ the sigmoid activation function to model the nonlinear variations in the appearance space and to constrain the value range. As for \textbf{morphology tokenizer} $\mathcal{E}_s,\mathcal{E}_q$, we simply use the normalization layer to standardize the output. Finally, the spatial, appearance, and morphological features are concatenated and fused via a two-layer MLP, yielding a unified 3DGS token that holistically captures attribute interdependencies for downstream tasks.


Inspired by UniGS's training strategy, our 3DGS token is prompted via cross-attention guidance from a pretrained point cloud model~\citep{zhou2023uni3d}. Specifically, position $\boldsymbol\mu$ and color $\boldsymbol c$ attributes are encoded as point cloud features, which are then refined through ViT blocks with cross-attention to integrate pretrained knowledge into the 3DGS token. Finally, the token is processed by a fully-connected layer (FC layer) to adjust its dimension to $d=512$ and output compact high-dimensional 3D features $F_\mathbb{G}^I\in\mathbb{R}^d$.

\subsection{Image-3D Alignment}

Conventional image-3D alignment methodologies ~\citep{xue2023ulip, zeng2023clip2, zhou2023uni3d, liu2023openshape, li2025unigs} predominantly focus on direct alignment between randomly selected single-view image and 3D features. However, this approach inherently compromises the perception capability of 3D features. Specifically, the alignment process constrained to a single perspective may lead to suboptimal matching performance for other viewpoints, as the cross-view consistency of 3D feature representations is not adequately preserved during the alignment procedure. While recent approaches attempt to leverage multiple text-image-3D triplets for a single 3D object for alignment (e.g., ULIP-2~\citep{xue2024ulip2}) or represent 3D objects directly using 2D multi-view images (e.g., Duoduo-CLIP~\citep{lee2024duoduo}), these methods inevitably compromise either computational efficiency (due to expanded training data requirements) or the benefits of concise explicit 3D representations. To address this issue, we propose a 3D-aware alignment scheme to bridge the gap between image features with 3D features.

As shown in Fig.~\ref{fig:align}(a), we leverage the multi-view diffusion~\citep{yang2024hunyuan3d} prior to elevate the image information from a single perspective to a 3D spatial dimension. For a single-view image $\mathbf{I}\in\mathbb{R}^{H\times W}$, we first generate $N$ multi-view images by the pretrained diffusion model:
\begin{equation}
    \mathcal D(\mathbf I,\Phi)=\{\mathbf I_0,\mathbf I_1,\cdots,\mathbf I_N\}, \textrm{where} \ \Phi=\{\phi_0,\phi_1,\cdots,\phi_N\},
\end{equation}
 with $\phi_i$ as the $i$-th preset camera angles. Afterwards, each image is fed into CLIP module~\citep{radford2021learningclip} to get the corresponding embedding:
\begin{equation}
F_{\mathbf{I}}=\mathrm{CLIP}(\mathbf{I})\in\mathbb R^d, \quad
    F_{\mathbf{I}_i}=\mathrm{CLIP}(\mathbf{I}_i)\in\mathbb R^d.
\end{equation}
 
We then use a perspective-aware cross-attention module for multi-view feature fusion:
\begin{equation}
    \mathrm{Attn}(Q,K,V)=\mathrm{Softmax}(\frac{QK^\top}{\sqrt{d}})V,
\end{equation}
where $Q=F_\mathbf{I}$, $K=V=[F_{\mathbf{I}_0}|F_{\mathbf{I}_1}|\cdots|F_{\mathbf{I}_N}]+\mathrm{PE}(\Phi)$, which is the concatenation of all single-view features. $\mathrm{PE}(\cdot)$ stands for sinusoidal positional encoding~\citep{vaswani2017attention}. The fused 3D-aware feature is:
\begin{equation}
    F_{\mathbb{I}}^{{mv}} = \mathrm{LayerNorm}(F_\mathbf{I}+\mathrm{Attn}(Q,K,V)) \in\mathbb{R}^{d}.
\end{equation}

During alignment, we replace the image feature $F_\mathbb{I}$ in Eq.~\ref{eq:loss} with $F_\mathbb{I}^{{mv}}$. In this way, 3DGS features will have multi-perspective perception capabilities, thereby enhancing the encoding ability in 3D space.

\subsection{Text-3D Alignment}
Although the current 3D features $F_{\mathbb G}^I$ has been aligned with multi-view image features, it is still not sufficient to produce good alignment with text modalities. In order to further narrow the gap between 3DGS and text modalities, we propose a feature projection module that aligns the latent space of 3DGS features and text features, thereby reducing the difficulty of feature alignment. Inspired by previous methods~\citep{li2023blip2, sirnam2024xformer, hadgi2025escaping}, we utilize a query transformer architecture for latent space projection. 

As shown in Fig.~\ref{fig:align}(b), we employ a set of learnable queries $F_q\in\mathbb{R}^{N_q\times d}$ as soft prompts to iteratively refine the 3D features through an $L$-layer transformer architecture. 
Each Transformer block consists of three modules: (1) self-attention for query refinement, (2) cross-attention to integrate 3D feature context, and (3) an MLP layer as the feed-forward network. These modules are connected via residual connections.






In detail, the transformer blocks process query token $F_q$ through $L$ sequential layers, where each layer $l\in[1,L]$ performs the following operations:
\begin{equation}
\begin{aligned}
F_q^l &= \mathrm{LayerNorm}\big(F_q^{l-1} + \mathrm{SelfAttn}(F_q^{l-1})\big), \\
F_q^l &= \mathrm{LayerNorm}\big(F_q^l + \mathrm{Attn}(F_q^l, F_\mathbb{G}^I, F_\mathbb{G}^I)\big), \\
F_q^l &= \mathrm{LayerNorm}\big(F_q^l + \mathrm{MLP}(F_q^l)\big),
\end{aligned}
\end{equation}
with initialization $F_q^0 = F_q$. Finally, the refined queries are flattened and passed through a pooling layer to produce the compact embedding $F_\mathbb{G}^T\in\mathbb{R}^d$ for text alignment.

In summary, the overall loss function for contrastive learning can be specialized to:
\begin{equation}
    \mathcal{L}=\lambda_\mathbb{T}\mathcal{L}(F_\mathbb{G}^T,F_\mathbb{T})+\lambda_\mathbb{I}\mathcal{L}(F_\mathbb{G}^I,F_\mathbb{I}^{{mv}}),
\end{equation}
which aims to align features preprocessed for different modalities and bridges the gap between them. 
Since our work focuses on 3D modality alignment and existing methods~\citep{sun2023evaclip, li2023blip2} have effectively addressed multi-modal tasks between text and images, we intentionally exclude the contrastive loss between text features and fused image features in our framework.

\section{Experiments}
\label{sec:exp}

\subsection{Experimental Setup}
\noindent\textbf{Datasets.}
We conduct our experiments on three public 3D datasets: Objaverse~\citep{deitke2023objaverse}, ABO~\citep{collins2022abo} and real-world indoor dataset SUN RGBD~\citep{song2015sun}, which contain 146k, 7.9k and 6.1k objects, respectively. As for multi-view images, we adopt the MVD-std model of Hunyuan3D-v1~\citep{yang2024hunyuan3d} to generate 6 views in canonical camera poses for each object before training. Please refer to \textit{supplementary materials} for more details.

\noindent\textbf{Implementation.} We use the Open-CLIP ViT-B-16 model~\citep{radford2021learningclip} as pre-aligned text-image model and Uni3D-S point cloud model~\citep{zhou2023uni3d} for 3DGS token guidance. We first train \method on Objaverse with the AdamW optimizer~\citep{adamw} and the learning rate of $1e^{-4}$ for 15 epochs for all downstream tasks. For ABO and SUN RGBD dataset, we quickly finetune the trained model for another 20 epochs to better fit the dataset. All features are set to the same dimension as $d=512$ for simplicity. Layer number of 3D-text projector is set to $L=6$. The balance factors $\lambda_\mathbb{T}$ and $\lambda_\mathbb{I}$ for contrastive learning are both set to 0.5, following the previous works~\citep{zhou2023uni3d, li2025unigs}. All models are trained on 4 A100 GPUs, and inference is performed on a single A100 GPU.

\noindent\textbf{Baselines.}
We compare our method with several state-of-the-art 3D multi-modal methods: CLIP$^2$~\citep{zeng2023clip2}, Uni3D~\citep{zhou2023uni3d}, UniGS~\citep{li2025unigs}. We directly inherit the experiment results reported in UniGS. We also reimplement ULIP-2~\citep{xue2024ulip2} and Duoduo-CLIP~\citep{lee2024duoduo} on the same data scales for classification tasks for fair comparisons. Please refer to \textit{supplementary materials} for more implementation details.

\subsection{Performance Comparison}

\noindent\textbf{Zero-shot Classification.} 
We evaluate the zero-shot classification performance of \method on the Objaverse-LVIS and ABO datasets, which contain 318 and 23 categories, respectively. Table~\ref{tab:classification} presents the Top-1, Top-3, and Top-5 classification accuracy results. The experimental results demonstrate that \method consistently outperforms all baseline methods on both datasets. Especially compared to recent 3DGS-based~\citep{li2025unigs} and multi-view based models~\citep{lee2024duoduo}, our method’s consistent superior performance demonstrates deeper exploration of 3DGS’s representational advantages in our design, leading to enhanced overall perception and cross-modal capabilities.
\begin{table}[htb]
    \centering\small
    \caption{Performances of different methods on zero-shot classification task. We report the average classification accuracy across all categories.}
    \begin{tabular}{l|ccc|ccc|c}
        \toprule 
        \multirow{2}{*}{Methods}  
        & \multicolumn{3}{c|}{\textit{Objaverse-LVIS}}
        &\multicolumn{3}{c|}{\textit{ABO}}
        & \multirow{2}{*}{3D Repr.}\\
        & Top-1 & Top-3 & Top-5 & Top-1 & Top-3 & Top-5 &\\
        \midrule
        \midrule
        CLIP$^2$ & 12.35 & 24.62 & 32.91 & 22.58 & 43.83 & 54.56 & Point Cloud \\
        ULIP-2 & 29.75 & 46.39 & 55.23 & - & - & - & Point Cloud \\
        Uni3D & 36.72 & 57.09 & 65.18 & 37.60 & 59.68 & 70.22 & Point Cloud \\
        UniGS & 37.64 & 57.62 & 65.57 & 52.33 & 70.27 & 79.38 & 3DGS \\
        Duoduo CLIP & 38.05 & 57.79 & 66.70 & 57.82 & 76.08 & 83.46  & Multi-view (non explicit) \\
        Ours & \textbf{41.76} &\textbf{62.68} & \textbf{69.15} & \textbf{61.70} & \textbf{83.16} & \textbf{89.79} & 3DGS \\
        \bottomrule
    \end{tabular}
    
    \label{tab:classification}
\end{table}

\noindent\textbf{Text-3D Retrieval.}
We select 1000 objects from Objaverse and ABO dataset for text-3D retrieval following the sampled test list provided by UniGS. Specifically, we simply calculate the cosine similarity of text embedding $F_\mathbb{T}$ and text-aligned 3DGS feature $F_\mathbb{G}^T$. Tab.~\ref{tab:combined-retrieval} shows the Top-1, Top-5 and Top-10 accuracy. Experimental results show that the proposed text projection module combined with multi-branch Gaussian tokenizer can effectively improve the correlation between 3D and text modalities.
\begin{table}[htb]
    \centering
    \small
    \setlength{\tabcolsep}{9pt}
    \caption{Performance comparison on text-3D and image-3D retrieval tasks. Reported metrics are average retrieval accuracy.}
    \begin{tabular}{l|ccc|ccc|c}
        \toprule
        \multirow{2}{*}{Methods} & 
        \multicolumn{3}{c|}{\textbf{Image-3D Retrieval}} & 
        \multicolumn{3}{c|}{\textbf{Text-3D Retrieval}} & 
        \multirow{2}{*}{3D Repr.} \\
        & Top-1 & Top-3 & Top-5 & Top-1 & Top-5 & Top-10 & \\
        \midrule
        \midrule
        \multicolumn{8}{c}{\textit{Objaverse-LVIS / Objaverse}} \\
        \midrule
        CLIP$^2$ & 28.83 & 51.43 & 63.57 & 7.40 & 22.20 & 32.50 & PointCloud \\
        Uni3D & 39.65 & 60.72 & 70.51 & 16.70 & 37.10 & 48.10 & Point Cloud \\
        UniGS & 41.78 & 62.50 & 72.24 & 21.00 & 39.80 & 53.50 & 3DGS \\
        Ours & \textbf{54.11} & \textbf{73.84} & \textbf{81.21} & \textbf{21.20} & \textbf{45.10} & \textbf{56.30} & 3DGS \\
        \midrule
        \midrule
        \multicolumn{8}{c}{\textit{ABO}} \\
        \midrule
        CLIP$^2$ & 15.29 & 31.74 & 42.74 & 7.09 & 24.34 & 38.94 & Point Cloud \\
        Uni3D & 18.25 & 35.26 & 45.29 & 10.29 & 29.21 & 43.67 & Point Cloud \\
        UniGS & 26.69 & 46.26 & 56.72 & 11.27 & 30.32 & 43.95 & 3DGS \\
        Ours & \textbf{66.15} & \textbf{73.99} & \textbf{85.27} & \textbf{15.87} & \textbf{40.17} & \textbf{53.07} & 3DGS \\
        \bottomrule
    \end{tabular}
    
    \label{tab:combined-retrieval}
\end{table}

\noindent\textbf{Image-3D Retrieval.}
We further evaluate the image-3D retrieval performances in each batch for Objaverse-LVIS and ABO dataset. The results shown in Tab.~\ref{tab:combined-retrieval} indicate that \method significantly surpasses all current baselines. We attribute it to the adjustment of image features and the integration of 3D information, thus overcoming the limitations of perspective issues between 3D-2D modal alignment.


\noindent\textbf{Few-shot Linear Probing.} 
To further evaluate the learning capabilities of our proposed model, we perform the few-shot linear probing following previous works~\citep{liu2023openshape, zhou2023uni3d}. Specifically, we freeze the 3DGS tokenizer and only train a linear classifier on few-shot class labels. We conduct few-shot linear probing on Objaverse-LVIS dataset with labeled training samples per class from 1, 2, 4, 8 to 16. We evaluate each model 10 times and report the average accuracy in Fig.~\ref{fig:linear-probing}. Notably, under the same amount of training data, \method outperforms the SoTA 3DGS-based method UniGS.

\noindent
\begin{minipage}[t]{0.4\linewidth}
\vspace{0pt}
\centering    
    \includegraphics[width=\linewidth]{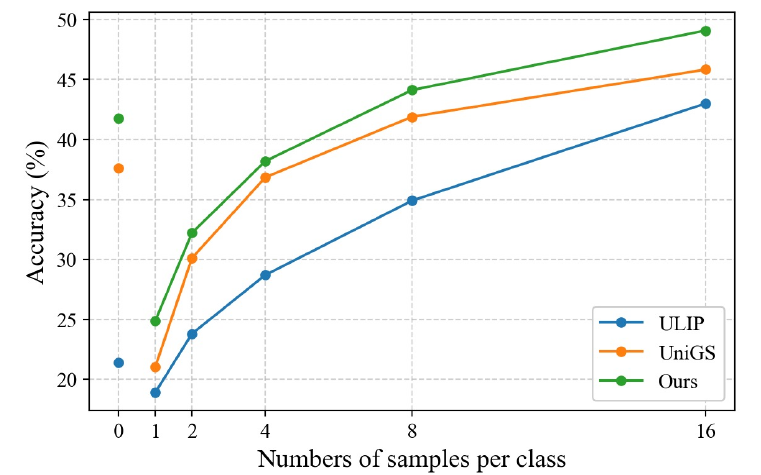}
    \captionof{figure}{Few-shot linear probing results on Objaverse-LVIS dataset. \textbf{0 number of samples per class} stands for zero-shot classification.}
    \label{fig:linear-probing}
\end{minipage}
\hfill
\begin{minipage}[t]{0.55\linewidth}
\vspace{0pt}
    \centering\small
    \begin{tabular}{l|c|c}
        \toprule 
        Methods & Mean Accuracy  & 3D Representation \\
        \midrule 
        \midrule 
        \multicolumn{3}{c}{\textit{SUN RGBD}} \\
        \midrule
        CLIP$^2$ &  41.39 & Point Cloud \\
        Uni3D &  61.72 & Point Cloud \\
        UniGS & 68.92   & 3DGS \\
        Ours & \textbf{76.46}  & 3DGS \\
        \bottomrule
    \end{tabular}
    \captionof{table}{Scene Recognition on SUN RGBD Dataset. We report the average recognition accuracy across all categories. Please refer to \textit{supplementary materials} for detailed results.}
    \label{tab:recognition}

\end{minipage}

\noindent\textbf{Open-world Scene Recognition.}
We follow UniGS~\citep{li2025unigs} to group the objects in SUN RGBD into 37 categories and take the Top-1 classification accuracy as the scene recognition accuracy. Tab.~\ref{tab:recognition} shows the average accuracy across all categories. The experimental results once again demonstrate the effectiveness of \method.

\begin{table}[htb]
\centering
\setlength{\tabcolsep}{5pt}\small
\caption{Ablation studies on each components. We report the Top 1. accuracy of each task on Objaverse dataset. We abbreviate multi-branch 3DGS tokenizer as \textbf{Tkn.}, using multi-view images as \textbf{MV.}, multi-view image fusion module as \textbf{MVF.}, and 3D-text projector as \textbf{TP.}. Moreover, we abbreviate classification as \textbf{Cl.}, text retrieval as \textbf{TR.}, and image retrieval as \textbf{IR.}.}
    \begin{tabular}{c|cccc|c|c|c}
    \toprule
     Exp. & Tkn. & MV. & MVF. & TP. & Cl. & TR. & IR. \\ \midrule
     1 & -    &  - & -  &  -   & 33.64 &  18.50        &     39.87        \\
     2 & \checkmark  & -   &  -   &  -   & 35.57 &  19.15        &  41.68           \\
     3 & \checkmark  & \checkmark   &  -   & -    & 37.28   & 19.16  & 46.19         \\
     4 & \checkmark  & \checkmark   &  \checkmark   & -    & 38.68  &  19.20       &    53.75         \\
     5 & \checkmark  & -   &  -   &  \checkmark   & 35.71  & 20.80        &     40.52        \\ 
     6 & -  & \checkmark   &  \checkmark   &  \checkmark   & 37.72 & 17.80         &   52.26          \\
     \midrule
     7 & \checkmark  & \checkmark   & \checkmark    & \checkmark  & \textbf{41.76} &  \textbf{21.20}   &  \textbf{54.11}           \\ \bottomrule
    \end{tabular}
    
    \label{tab:ablation}
\end{table}

\subsection{Ablation Studies}

\noindent\textbf{Ablations on 3DGS Tokenizer.}
We first validate the necessity of the multi-branch 3DGS tokenizer which disentangles the Gaussian attributes. Tab.~\ref{tab:ablation} shows the Top-1 accuracy of each task on the Objaverse dataset with different experiment setups. 
As shown in Tab.~\ref{tab:ablation}, the naive UniGS (Exp1) performs less well compared to Exp2, with the 3DGS tokenizer consistently improving accuracy in all tasks. Further analysis (Exp6 vs. Exp7) confirms its role as a crucial 3D context extractor during feature alignment. Removing this component causes significant performance degradation, demonstrating its fundamental importance. These results show that our tokenizer effectively resolves the problem of 3D context abstraction.

\noindent\textbf{Ablations on Alignment Strategy.}
We validate our modality alignment approach through extensive experiments on image and text alignment strategies.  Key findings include:
\begin{itemize}
    \item The \textbf{multi-view images feature fusion} (Exp2 vs. Exp4) strategy effectively bridges the image-3D modality gap, significantly improving image retrieval and zero-shot classification.
    \item \textbf{Removing the fusion module} (Exp4 vs. Exp3) leads to noticeable performance degradation, confirming that simply using more training triplets incurs computational costs and doesn't have sufficient performance gains.
    \item The \textbf{3D-text projector} (Exp2 vs. Exp5) successfully aligns 3DGS features with textual representations.
\end{itemize}
The integrated system (Exp7), combining multi-branch 3DGS tokenizer, 3D-aware image fusion, and 3D-text projector, achieves optimal performance, demonstrating the complementary roles of these modules in cross-modal alignment.


\noindent\textbf{More Ablation Studies and Discussions. } We also provide detailed discussions and experiments on \textbf{guidance strategy}, \textbf{scaling-up}, and \textbf{generalization}. Please check the \textit{supplementary materials} for more.
\section{Conclusion and Discussions}
This paper presents \method, a framework that establishes robust alignment among text, image, and 3DGS modalities by addressing three challenges in cross-modal 3D understanding. First, our multi-branch 3DGS tokenizer scheme systematically decouples 3DGS intrinsic attributes, enabling better feature representation of 3D structures. Second, the diffusion-enhanced multi-view fusion mechanism resolves the inherent limitations of single-view observations, ensuring spatial coherence across diverse perspectives and significantly improving the image-3D alignment. Third, the 3D-text projection module effectively bridges the modality gap between continuous 3D feature spaces and discrete text embeddings. Extensive experiments demonstrate state-of-the-art performance in multiple downstream tasks. Our work establishes a new paradigm for leveraging compact latent 3D token in vision-language applications, which further stimulates the potential of 3DGS as an advanced 3D representation in cross-modal understanding.

\noindent\textbf{Limitations and Future Works.}
\method has been validated to achieve good results across different datasets and tasks. However, there are two limitations that merit attention. (1) Generalization ability: Although we conduct experiments on Gaussians from various sources and verified the scalability, there may still be scenarios where our method performs degraded on occluded multi objects or real outdoor scenes.
(2) Text label dependency: Currently, the quality of text-3D alignment fundamentally relies on training labels. Current benchmarks primarily utilize LLM-generated annotations, which can introduce bias. Future work could explore hybrid supervision paradigms that combine the efficiency of LLMs with expert data label curation for better fine-grained text retrieval.

\bibliographystyle{iclr2026_conference}
\bibliography{main}

\clearpage
\appendix
\appendix

\section{Implementation Details}
\subsection{Dataset Details}
For each dataset (i.e., Objaverse~\citep{deitke2023objaverse}, ABO~\citep{collins2022abo} and SUN RGBD~\citep{song2015sun}), we prepare the 3DGS models following methods by UniGS~\citep{li2025unigs}, and sample 1024 Gaussian primitives for each object except for SUN RGBD dataset. As for the text annotations and of each object, we use InternLM-Xcomposer~\citep{dong2024internlm} to auto-generate the text prompt. To ensure the fairness of comparison, we follow UniGS~\citep{li2025unigs} for datasets splitting and evalutaion setup. Specifically, we use the same test sets for all dataset and Objaverse-LVIS for evaluation.

\subsection{Training Details} 
We first train \method for 15 epochs on Objaverse dataset~\citep{deitke2023objaverse} with 100k objects and evaluate on Objaverse-LVIS dataset. The batch size is 24 for training and 80 for testing. For ABO~\citep{collins2022abo} and SUN RGBD~\citep{song2015sun} dataset, we quickly finetune the trained model for another 20 epochs to better fit the dataset.

\subsection{Reimplementation Details of Baselines}
\paragraph{UniGS.}
We inherit the test results of UniGS~\citep{li2025unigs} from the original UniGS paper. Notice that they \textbf{DO NOT} release the pretrained checkpoints, and the re-implemented results are slightly downgraded compared to the public results, thus we strictly follow their reported results for fair comparison. The guidance model is Uni3D-S, which is the same used for \method. Overall, the performance of UniGS generally surpasses other baseline methods. 
\paragraph{ULIP-2 and Duoduo CLIP.}
We follow the training instruction of ULIP-2~\citep{xue2024ulip2} and Duoduo CLIP~\citep{lee2024duoduo}. As for multi-view images number selected for training, we keep it the same as \method, i.e., 6 images for each object. Moreover, we use the sampled dataset of Objaverse provided by UniGS to train the model instead of using ensemble dataset for fair comparison. It is worth noting that the training data of ULIP-2 has increased by 6 times directly, resulting in a sharp increase in training time and memory overhead.

\paragraph{Others.}
We inherit the test results of CLIP$^2$~\citep{zeng2023clip2} and Uni3D~\citep{zhou2023uni3d} from the UniGS paper. As for few-shot linear probing result of ULIP~\citep{xue2023ulip} shows in the main text, we follow the result reported in Uni3D~\citep{zhou2023uni3d}.

\section{More Experimental Results}

\subsection{Detailed Results of Scene Recognition}
We report the scene recognition accuracy of 10 typical classes in SUN RGBD dataset in Tab.~\ref{tab:detail-recognition} for detailed comparison with UniGS. 

\begin{table*}[h]
    \centering\scriptsize
     \caption{Recognition results on SUN RGBD dataset. Notice that the average accuracy is across all 37 categories.}
    \begin{tabular}{l|cccccccccc|c}
        \toprule 
        Methods & Bed & Bookshelf & Chair & Desk & Sofa & Table & Toilet & Bathtub & Dresser & Nightstand & Average \\
        \midrule 
        UniGS & 80.57 & 76.49 & 88.65 & 65.45 & 88.61 & 74.23 & \textbf{93.45} & 92.06 & 57.64 & 60.16 & 68.92 \\
        Ours & \textbf{93.87} & \textbf{79.33} & \textbf{91.62} & \textbf{72.62} & \textbf{90.56} & \textbf{77.50} & 92.82 & \textbf{94.67} & \textbf{67.50} & \textbf{62.15} & \textbf{76.46} \\
        \bottomrule
    \end{tabular}
   
    \label{tab:detail-recognition}
\end{table*}

\subsection{Comprehensive Comparisons with More Baseline Methods}
\label{sec:rebuttal}
We conduct additional baseline comparison on OpenShape~\citep{liu2023openshape}, TAMM~\citep{zhang2024tamm}, MixCon3D~\citep{gao2023mixcon3d}, Recon~\citep{qi2023recon} and ReCon++~\citep{qi2024shapellm}. To ensure fair comparison, we reimplement these baselines by training on Objaverse 100k dataset, and also scaling up our \method to train on Objaverse 800k dataset without Objaverse-LVIS.

Notice that since most of the baselines leverage an ensembled point cloud dataset (including Objaverse-LVIS evaluation dataset) for training, which has about 876k instances with 10000 points per object, while the results reported in our paper are conducted on only 100k training instances with 1024 points per object.
The comprehensive results shown in Tab.~\ref{tab:rebuttal-A} and Tab.~\ref{tab:rebuttal-B} reveal the superior performance of our proposed method, demonstrating the effectiveness of 3DGS representations and the proposed tokenization and alignment strategies.

\begin{table}[htb]
    \caption{Comparison of object classification task on Objaverse-LVIS datset.}
    \centering
    \begin{tabular}{l|c|c|c|c c c}
        \toprule 
        Methods & Base Model & Training Dataset & GS Points & Top-1 & Top-3 & Top-5  \\
        \midrule 
        \midrule
        \multicolumn{7}{c}{\textit{Objaverse-LVIS}} \\
        \midrule
        Recon   & PointBERT & \multirow{8}{*}{Objaverse-100k} & 10000 & 25.1 & 45.6 & 50.1 \\
        TAMM    & PointBERT &                                   & 10000 & 22.7 & 40.1 & 48.5 \\
        MixCon3D& PointBERT &                                   & 10000 & 32.3 & 52.5 & 61.5 \\
        ReCon++ & ViT-S     &                                   & 10000 & 26.9 & 47.5 & 53.6 \\
        ReCon++ & ViT-B     &                                   & 10000 & 31.0 & 50.5 & 55.2 \\
        Ours-S  & Uni3D-S   &                                   & 1024  & 41.8 & 62.7 & 69.1 \\ 
        Ours-S  & Uni3D-S   &                                   & 10000 & \textbf{46.7} & \textbf{68.6} & 74.5 \\
        Ours-B  & Uni3D-B   &                                   & 1024  & 46.6 & 67.5 & \textbf{75.4} \\
        \bottomrule
    \end{tabular}
    \label{tab:rebuttal-A}
\end{table}

\begin{table}[htb]
    \caption{Comparison of object classification task on Objaverse-LVIS datset with larger training dataset.}
    \centering
    \begin{tabular}{l|c|c|c|c c c}
        \toprule 
        Methods & Base Model & Training Dataset & GS Points & Top-1 & Top-3 & Top-5  \\
        \midrule 
        \midrule
        \multicolumn{7}{c}{\textit{Objaverse-LVIS}} \\
        \midrule
        OpenShape & PointBERT & Ensembled w/o LVIS & 10000 & 39.1 & 60.8 & 68.9 \\
        TAMM & PointBERT & Ensembled w/o LVIS & 10000 & 42.0 & 63.6 & 71.7 \\
        MixCon3D & PointBERT & Ensembled w/o LVIS & 10000 & 47.5 & 69.6 & 76.2 \\
        ReCon++ & ViT-B & Ensembled w/o LVIS & 10000 & 49.6 & 70.2 & 78.4 \\
        UniGS & Uni3D-S & Objaverse-800k w/o LVIS & 1024 & 48.8 & 69.1 & 76.9 \\
        Ours & Uni3D-S & Objaverse-800k w/o LVIS & 1024 & \textbf{50.1} & \textbf{73.6} & \textbf{79.6} \\
        \bottomrule
    \end{tabular}
    \label{tab:rebuttal-B}
\end{table}

\section{Ablations on Guidance}
Compared to raw point cloud, 3DGS has richer structural priors, since each Gaussian contains position, color, scale, rotation and opacity, which enables explicit modeling of local geometry and view-dependent appearance. As we proposed in the paper, the position and color attribute of 3DGS is similar to point cloud, thus the spatial tokenizer could directly benefit from the pre-trained point cloud encoder (e.g., Uni3D~\citep{zhou2023uni3d} with ViT~\citep{dosovitskiy2020vit}). Besides, the pre-trained model is trained on much larger point cloud datasets, thus its generalizability could be inherited and make the training process easier. Based on this design, our proposed 3DGS tokenizer fully utilizes other attributes of GS to better perform multimodal alignment. Furthermore, we believe the choice of 3D representation has long-term implications for generative and downstream tasks. While point clouds are simpler, 3DGS bridges the gap between discrete 3D elements and continuous radiance fields — offering a promising direction for future 3D-aware generation and understanding systems.
We add additional experiments on training 3DGS encoder without pretrained Uni3D point cloud model guidance. As shown in Tab.~\ref{tab:guide}, leveraging pretrained model has significant advantages. 

\begin{table}[htb]
    \centering
    \caption{Performances of different methods on zero-shot classification task. We report the average classification accuracy across all categories.}
    \begin{tabular}{l|ccc}
        \toprule 
        Methods & Top-1 & Top-3 & Top-5  \\
        \midrule 
        \midrule 
        \multicolumn{4}{c}{\textit{Objaverse-LVIS}} \\
        \midrule
        Ours w/o pretrained & 27.93 & 47.74 & 57.08 \\
        Ours & \textbf{41.76} & \textbf{62.68} & \textbf{69.15} \\
        \bottomrule
    \end{tabular}
    
    \label{tab:guide}
\end{table}

\section{Ablations on Multi-view Diffusion Module}
\label{sec:mvd}
We also analysis the impact of using different multi-view diffusion models or ground-truth rendered images. We select 6 multi-view images for Objaverse dataset following several methods: (1) Hunyuan3D-v1-lite~\citep{yang2024hunyuan3d}, (2) Hunyuan3D-v1-std, (3) rendered ground-truth images given by Zero-1-to-3~\citep{liu2023zero}. As shown in Tab.~\ref{tab:mvd}, the results are slightly different from each other depending on the performance of chosen diffusion model.

\begin{table}[htb]
    \centering
    \caption{Performances of our method training on different source of multi-view images on zero-shot classification task. We report the average classification accuracy across all categories.}
    \begin{tabular}{l|ccc}
        \toprule 
        Methods & Top-1 & Top-3 & Top-5  \\
        \midrule 
        \midrule 
        \multicolumn{4}{c}{\textit{Objaverse-LVIS}} \\
        \midrule
        Ours w/ Hunyuan3D-v1-lite & 39.98 & 61.13 & 68.42 \\
        Ours w/ Hunyuan3D-v1-std & \textbf{41.76} & \textbf{62.68} & 69.15 \\
        Ours w/ GT & 41.03 & 62.31 & \textbf{69.78} \\
        \bottomrule
    \end{tabular}
    
    \label{tab:mvd}
\end{table}

\section{Ablations on Multi-view Image Numbers}
To estimate the impact of the number of multi-view images involved in \method, we conduct an additional ablation study on different numbers of multi-view images as shown in Tab.~\ref{tab:mv-numbers}. xperimental results have shown that selecting more multi-view images does not necessarily mean better multimodal alignment. We hypothesize that this is due to over-alignment between the image and 3DGS modalities when too many views are used, causing the fused visual features to become overly specific and thus harder to align with the more abstract text representations. In addition, as the number of perspectives increases, the requirement for the authenticity of multiple perspectives will also increase, and the number of model parameters will also multiply. Experimental data shows that selecting 6 multi view images can achieve a balance.

\begin{table}[htb]
    \centering
    \caption{Performances of our method training with different numbers of multi-view images.}
    \begin{tabular}{l|c|c}
        \toprule 
        \# MV Images & Classification Top-1 & Image Retrieval Top-1  \\
        \midrule 
        \midrule 
        \multicolumn{3}{c}{\textit{Objaverse-LVIS}} \\
        \midrule
        1 & 35.71 & 40.52  \\
        3 & 39.61 & 46.79 \\
        6 & \textbf{41.76} & 52.11 \\
        12 & 40.52 & \textbf{52.37} \\
        \bottomrule
    \end{tabular}
    
    \label{tab:mv-numbers}
\end{table}

\section{Runtime and Cost Analysis}
We further compare the inference time cost with baseline methods on Objaverse-LVIS dataset. All time tests are conducted on a single A100 GPU. As shown in Tab.~\ref{tab:time}, our method strike a balance between time cost and accuracy. Notice that our disentangled encoder does not introduce too much additional overhead while achieving high performance surpassing compared to UniGS. Compared to methods based on multi-view representation (i.e., Duoduo CLIP~\citep{lee2024duoduo}), our inference time has a significant advantage. As for multi-view generation part, we take it as the training overhead, as Hunyuan3D-v1-lite takes 800ms to generate 6 views per image. Yet as we've discussed in Sec.~\ref{sec:mvd}, when using ground-truth multi-view images provided by original Zero-1-to-3 dataset, there is no significant training overhead involved. 

Regarding the model complexity, the GFLOPs of 3DGS Tokenizer is 88.2M, which takes about 80\% of the total FLOPs our model. The image fusion module takes 1M for cross attention only, and 3D-text projector takes 24M for multi-layer transformer. The maximum GPU memory of TIGaussian is 32G during inference, which can be easily perfomed on a single RTX4090 GPU.

\begin{table}[htb]
    \centering
    \caption{Comparisons of inference computational cost on Objaverse-LVIS dataset.}
    \begin{tabular}{l|c|c}
        \toprule
        Methods & Inference Time (ms) & Top-1 Acc. \\
        \midrule 
        \midrule 
        \multicolumn{3}{c}{\textit{Objaverse-LVIS}} \\
        \midrule
        Uni3D & 97.96 & 36.72 \\
        UniGS & 185.75 & 37.64 \\
        Duoduo CLIP & 235.79 & 38.05 \\
        Ours & 192.86 & 41.76 \\
        \bottomrule
    \end{tabular}
    
    \label{tab:time}
\end{table}

\section{Ablations on Text Captions}
In the main experiments, we use the InternLM-Xcomposer generated text prompt for text modal alignment. In order to further discuss the impact of text label selections, we conduct additional experiments on text label generated by Cap3D, an advanced text label tools which has been evaluated by human experts. As shown in Tab.~\ref{tab:text-prompt}, the results are slightly different. Shorter text prompts are more likely to have better classification result.

\begin{table}[htb]
    \caption{Performances of our method training on different source of text prompt on zero-shot classification task. We report the average classification accuracy across all categories.}
    \centering
    \begin{tabular}{l|ccc}
        \toprule 
        Methods & Top-1 & Top-3 & Top-5  \\
        \midrule 
        \midrule 
        \multicolumn{4}{c}{\textit{Objaverse-LVIS}} \\
        \midrule
        Ours w/ InternLM & \textbf{41.76} & \textbf{62.68} & \textbf{69.15} \\
        Ours w/ Cap3D & 40.12 & 60.35 & 67.82 \\
        \bottomrule
    \end{tabular}
    
    \label{tab:text-prompt}
\end{table}

\section{Analysis of Generalization Ability}
\begin{figure}[t]
    \centering
    \includegraphics[width=0.7\linewidth]{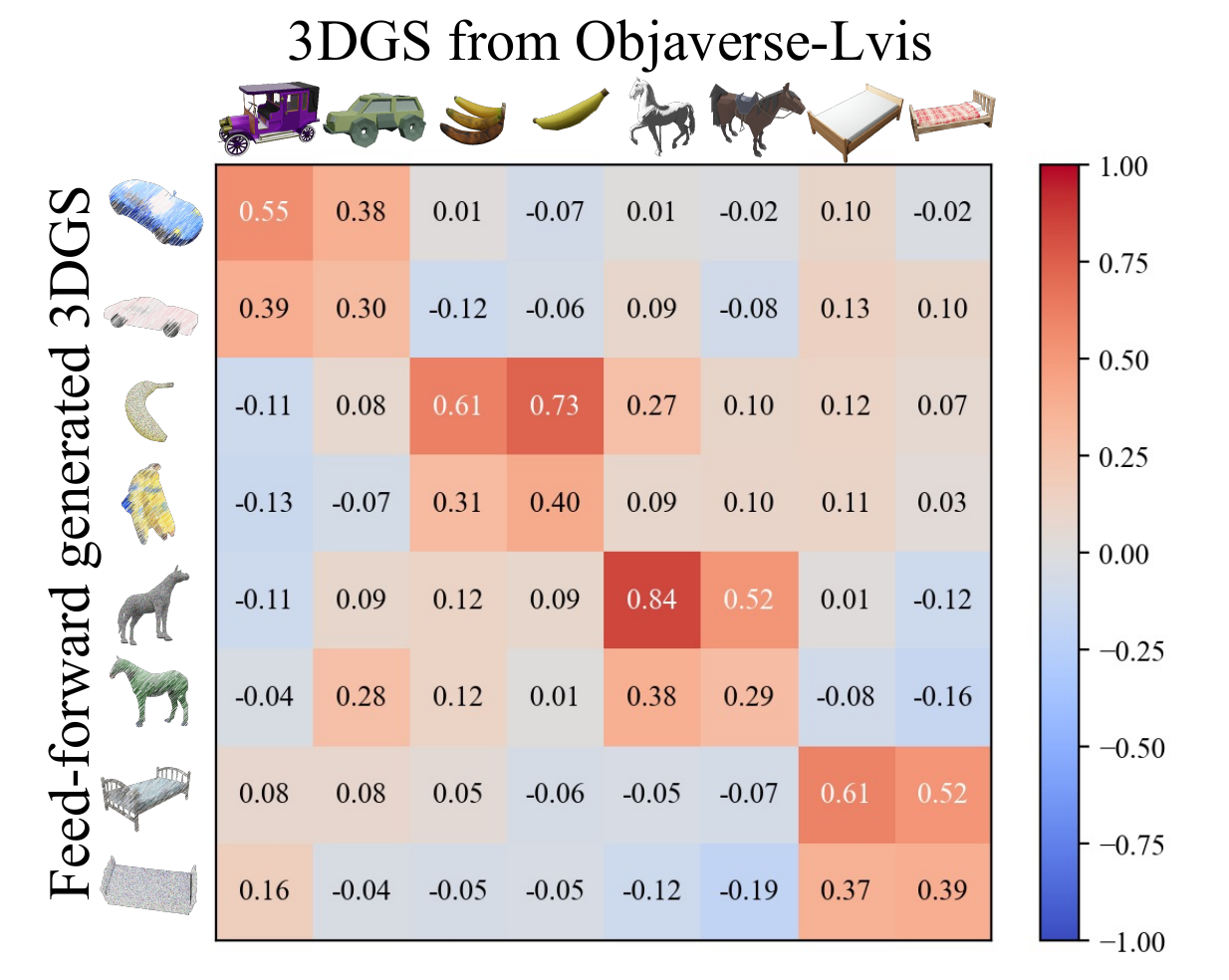}
    \caption{3DGS feature similarity matrix between Objaverse objects and feed-forward generated 3DGS objects. We select 4 categories from Objaverse-LVIS (i.e., cars, bananas, horses and beds), and compute the cosine similarity of the 3D features of original 3DGS (top) and generated 3DGS (left). The diagonal form of the heatmap proves that our 3D features provide relatively high similarity of objects from the same category, despite their different data sources.}
    \label{fig:generalization}
\end{figure}

To demonstrate the generalizability of our proposed method as a 3DGS encoder, we present similarity heatmaps of 3D embeddings from 3DGS objects of various sources. 
Fig.~\ref{fig:generalization} shows the correlation between original 3DGS objects and the generated 3DGS objects by feed-forward reconstruction method AnySplat~\citep{jiang2025anysplat}, which proves that our method is insensitive to the data source of 3DGS and has good generalization ability. Please refer to \textit{supplementary materials} for details. 

\begin{figure}[h]
    \centering
    \includegraphics[width=\linewidth]{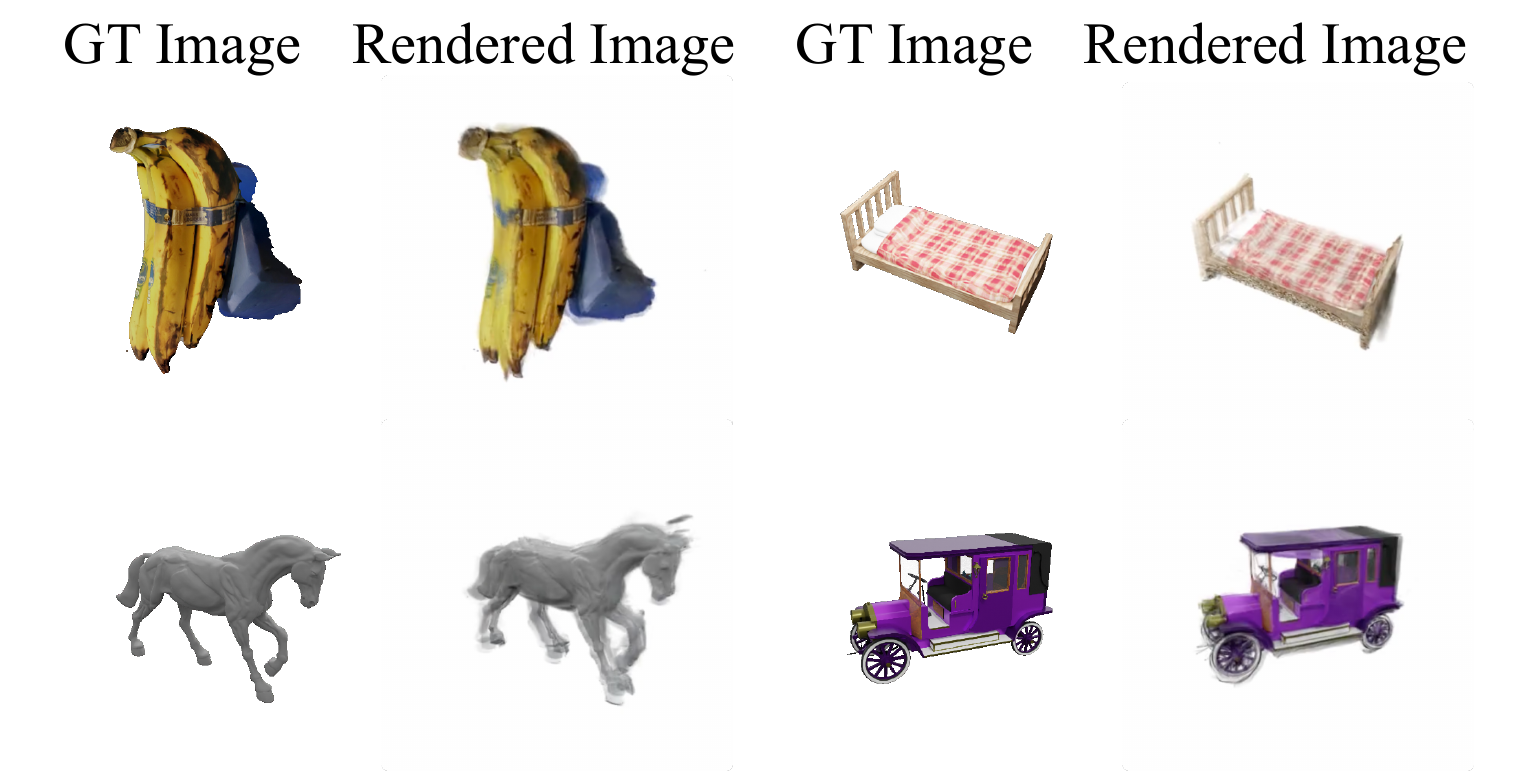}
    \caption{Comparison between the GT images and the rendered images of the 3DGS generated by AnySplat~\citep{jiang2025anysplat}.}
    \label{fig:splat}
\end{figure}

As for implementation, we leverage the VGGT-based~\citep{wang2025vggt} feed-forward 3DGS generation method AnySplat~\citep{jiang2025anysplat} to generate several 3DGS objects from the Objaverse-LVIS dataset. Specifically, we randomly select only 3 views of each object and directly get the 3DGS representation results. Fig.~\ref{fig:splat} shows some visualization results of the rendered image of the generated 3DGS.

\section{Model Scaling Up}

\begin{figure*}[t]
    \centering
    \includegraphics[width=0.9\linewidth]{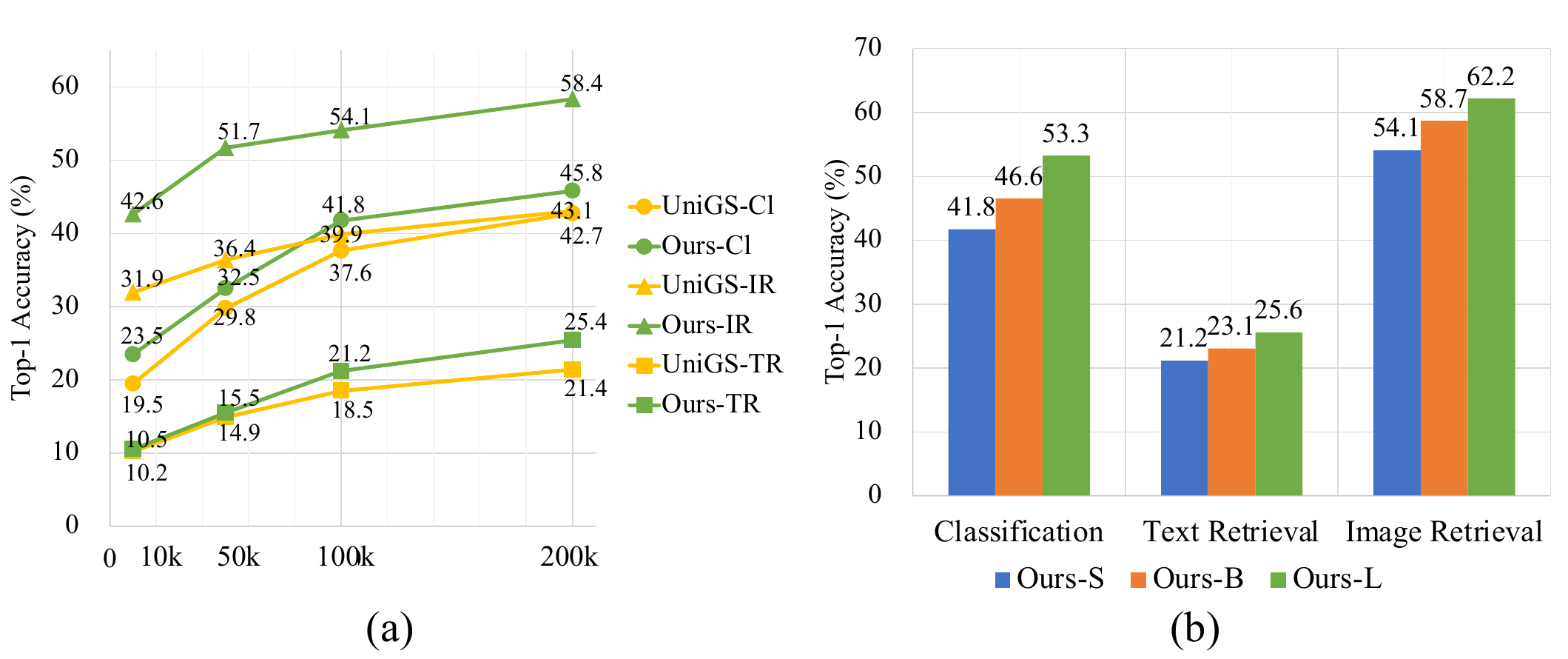}
    \caption{Analysis on scaling up the model. (a) Top-1 accuracy on Objaverse dataset with different scales of training data. We abbreviate classification as \textbf{Cl}, image retrieval as \textbf{IR} and text retrieval as \textbf{TR}. (b) Top-1 accuracy on Objaverse dataset with different backbone setups.}
    \label{fig:scaling-up}
\end{figure*}

We test the performances results of the proposed method and the baseline method on different scales of training data. As shown in Fig.~\ref{fig:scaling-up}(a), scaling up the training data of \method can effectively improve the performance on various downstream tasks. Meanwhile, our method surpasses the baseline method and demonstrates effective improvement in training with higher data volumes. In order to further demonstrate the upper limit of our proposed method, we used pretrained point cloud models with different parameter numbers (i.e., Uni3D-S, Uni3D-B and Uni3D-L) to train our encoder, result in Ours-S, Ours-B and Ours-L, respectively. Fig.~\ref{fig:scaling-up}(b) shows Top-1 performances of our method using different backbone. 

As for the detailed comparison of scaling up on training dataset and sampled 3DGS points, please refer to Tab.~\ref{tab:rebuttal-A} and Tab.~\ref{tab:rebuttal-B} in Sec.~\ref{sec:rebuttal}.

\end{document}